\documentclass[letterpaper,twocolumn,fleqn]{article}
\usepackage{spconf,amsmath,graphicx}
\usepackage{epsfig}
\usepackage{amssymb}
\usepackage{subfigure}
\usepackage{cite}
\usepackage{multirow}
\usepackage{epstopdf}
\usepackage{graphicx}
\usepackage{amsmath}
\usepackage{color, soul}
\usepackage[style=base]{caption}
\usepackage{lineno}
\usepackage{comment}
\usepackage{array}  
\usepackage{siunitx}
\usepackage{mathtools}
\usepackage{calrsfs}
\usepackage{eucal}
\usepackage{bm}
\newcolumntype{d}{>{\displaystyle}c}
\usepackage{url}
\usepackage{hyperref}
\hypersetup{
	colorlinks=true,       % false: boxed links; true: colored links
	linkcolor=red,          % color of internal links (change box color with linkbordercolor)
	citecolor=green,        % color of links to bibliography
	filecolor=magenta,      % color of file links
	urlcolor=cyan           % color of external links
}
\usepackage{enumitem}
\usepackage{pgfplots}
\usepackage{array}
\newcolumntype{L}[1]{>{\raggedright\let\newline\\\arraybackslash\hspace{0pt}}m{#1}}
\newcolumntype{C}[1]{>{\centering\let\newline\\\arraybackslash\hspace{0pt}}m{#1}}
\newcolumntype{R}[1]{>{\raggedleft\let\newline\\\arraybackslash\hspace{0pt}}m{#1}}
\title{Kalman Filter Based Multiple Person Head Tracking}

\name{
	\begin{tabular}{c}
		Mohib Ullah$^1$, Maqsood Mahmud$^2$, Habib Ullah$^3$, Kashif Ahmad$^4$,\\ Ali Shariq Imran$^1$, Faouzi Alaya Cheikh$^1$ 
	\end{tabular}
}

\address{
	\begin{tabular}{c}
		$^1$ Department of Computer Science (IDI), Norwegian University of Science and Technology, Norway. \\
		$^2$ Department of Management Information System, \\ University of Imam Abdulrahman Bin Faisal, Dammam, Saudi Arabia. \\ 
		$^3$ College of Computer Science and Engineering, University of Ha'il, Saudi Arabia. \\ 
		$^4$ College of Science and Engineering, Hamad Bin Khalifa University (HBKU), Doha, Qatar. 	
	\end{tabular}
}

\begin{document}
%\ninept
%
\maketitle

\begin{abstract}

\iffalse
In tracking-by-detection paradigm for multi-target tracking, target association is modeled as an optimization problem that is usually solved through network flow formulation. In this paper, we proposed combinatorial optimization formulation and used a bipartite graph matching for associating the targets in the consecutive frames. Usually, the target of interest is represented in a bounding box and track the whole box as a single entity. However, in the case of humans, the body goes through complex articulation and occlusion that severely deteriorate the tracking performance. To partially tackle the problem of occlusion, we argue that tracking the rigid body organ could lead to better tracking performance compared to the whole body tracking. Based on this assumption, we generated the target hypothesis of only the spatial locations of person's heads in every frame. After the localization of head location, a constant velocity motion model is used for the temporal evolution of the targets in the visual scene. Qualitative results are evaluated on four challenging video surveillance dataset and promising results has been achieved.          
\fi

For multi-target tracking, target representation plays a crucial rule in performance. State-of-the-art approaches rely on the deep learning-based visual representation that gives an optimal performance at the cost of high computational complexity. In this paper, we come up with a simple yet effective target representation for human tracking. Our inspiration comes from the fact that the human body goes through severe deformation and inter/intra occlusion over the passage of time. So, instead of tracking the whole body part, a relative rigid organ tracking is selected for tracking the human over an extended period of time. Hence, we followed the tracking-by-detection paradigm and generated the target hypothesis of only the spatial locations of heads in every frame. After the localization of head location, a Kalman filter with a constant velocity motion model is instantiated for each target that follows the temporal evolution of the targets in the scene. For associating the targets in the consecutive frames, combinatorial optimization is used that associates the corresponding targets in a greedy fashion. Qualitative results are evaluated on four challenging video surveillance dataset and promising results has been achieved.

\iffalse
For tracking multiple targets in a scene, the most common approach is to represent the target in a bounding box and track the whole box as a single entity. However, in the case of humans, the body goes through complex articulation and occlusion that severely deteriorate the tracking performance. In this paper, we argue that instead of tracking the whole body of a target, if we focus on a relatively rigid body organ, better tracking results can be achieved. Based on this assumption, we followed the tracking-by-detection paradigm and generated the target hypothesis of only the spatial locations of heads in every frame. After the localization of head location, a constant velocity motion model is used for the temporal evolution of the targets in the visual scene. For associating the targets in the consecutive frames, combinatorial optimization is used that associates the corresponding targets in a greedy fashion. Qualitative results are evaluated on four challenging video surveillance dataset and promising results has been achieved.

\fi

\end{abstract}
%

%------------------------------------------------------------------------------------------------------------------%
\section{Introduction}
\label{sec:intro}

\iffalse
Multi-feature-based crowd video modeling for visual event detection
\cite{ullah2020multi}

Facial Emotion Recognition Using Hybrid Features
\cite{alreshidi2020facial}

Two Stream Model for Crowd Video Classification
\cite{ullah2019two}

\cite{ullah2019stacked}

\fi

One of the primary tasks of machine learning is to enable computers to learn from the data and automatically do thoughtful predictions. Such capabilities has applications in airline scheduling \cite{bazargan2016airline}, crowd modeling \cite{ullah2020multi}, and face recognition based fraud detection \cite{alreshidi2020facial}. For the visual data, it helps to analyze and classify a visual scene \cite{ullah2019two, owens2018audio}. In the realm of visual scene analysis, multi-target tracking is one of the most important low-level computer vision problems that provides a backbone to many high level tasks like autonomous driving \cite{khan2019survey, li2019generic}, action recognition \cite{ullah2017human, wang2018generative, ullah2019stacked}, behavior analysis \cite{marsden2017resnetcrowd, ullah2016crowd}, anomaly detection\cite{wang2018generative, ullah2019hybrid, ullah2018anomalous}, crowd management \cite{ullah2017density, ullah2017density}, and sports players analysis \cite{thomas2017computer, khan2019person}, to name a few. Even though tracking on its own is a low-level computer vision problem, intrinsically, it consists of other low-level tasks like object segmentation\cite{ullah2018pednet}, object detection\cite{khan2019disam}, and motion modeling\cite{basalamah2019scale}. With the advancement in object detection algorithms \cite{redmon2016you, khan2019dimension}, the tracking-by-detection paradigm becomes the most suitable for tracking multiple objects in a visual scene. However, the biggest question that arises is which object part to track or track the whole object mass. Until now, almost all the tracking algorithms use the whole body detection \cite{voigtlaender2019mots, ullah2019single, schulter2017deep, ullah2017hierarchical, milan2014continuous, sun2019deep, ullah2018directed, bae2017confidence, ullah2016hog, son2017multi, chu2017online, ullah2018deep}. For example, Milan et al. \cite{milan2014continuous} proposed a highly non-convex cost function for multi-target tracking where different components like appearance, detection, motion, target mutual osculation, etc. are combined in a weighted average function. A gradient descent based optimization is used to optimize the cost and transdimentional jumps are used to avoid the local minima. Ullah et al. \cite{ullah2017hierarchical} proposed a bag of Bayesian filters to track multiple targets in the scene. Additionally, sparse coded deep features are incorporated to model the appearance of the targets. Schulter et al. \cite{schulter2017deep} modeled multi-target tracking as a network flow graph. Instead of calculating the edge cost of the graph manually through hand-crafted features, they learned the edge of the graph through back propagation. Similarly, Ullah et al. \cite{ullah2018deep} also generated a directed acyclic graph for the multi-target tracking but used deep features for calculating the edge cost of the graph. Dehghan et al. \cite{dehghan2015gmmcp} formulated the tracking problem as a multi clique problem. Initially, a graph is generated from a batch of frames and later each target trajectory is found as the maximum clique of the graph. Chu et al. \cite{chu2017online} proposed a spatial-temporal attention mechanism for occulation handling and the interaction among different targets. They extract features from the different layers of CNN for modeling the appearance of the target. Compare to that, \cite{ullah2019single} proposed a Siamese neural network for modeling the appearance of targets and establishing the association between two targets. 

One of the common attributes among all the tracking techniques is that they represent the whole human body as a rectangular rigid object, even though the human body goes through severe articulations. Compared to the standard approaches, in this paper, we focused on the relatively rigid organ of the human body i.e. head of a person and first generated the head hypothesis in all the frames and then used Combinatorial optimization to associate the target head in the consecutive frames. The organization of the paper is the following: In section \ref{PA}, the proposed approach is briefly explained. The tracker details are given in section \ref{tracker}. The qualitative and quantitative results are given in section \ref{EX} and section \ref{CON} concludes the paper with future directions.

\begin{figure}%[htp]
	\centering
	\includegraphics[scale=0.38]{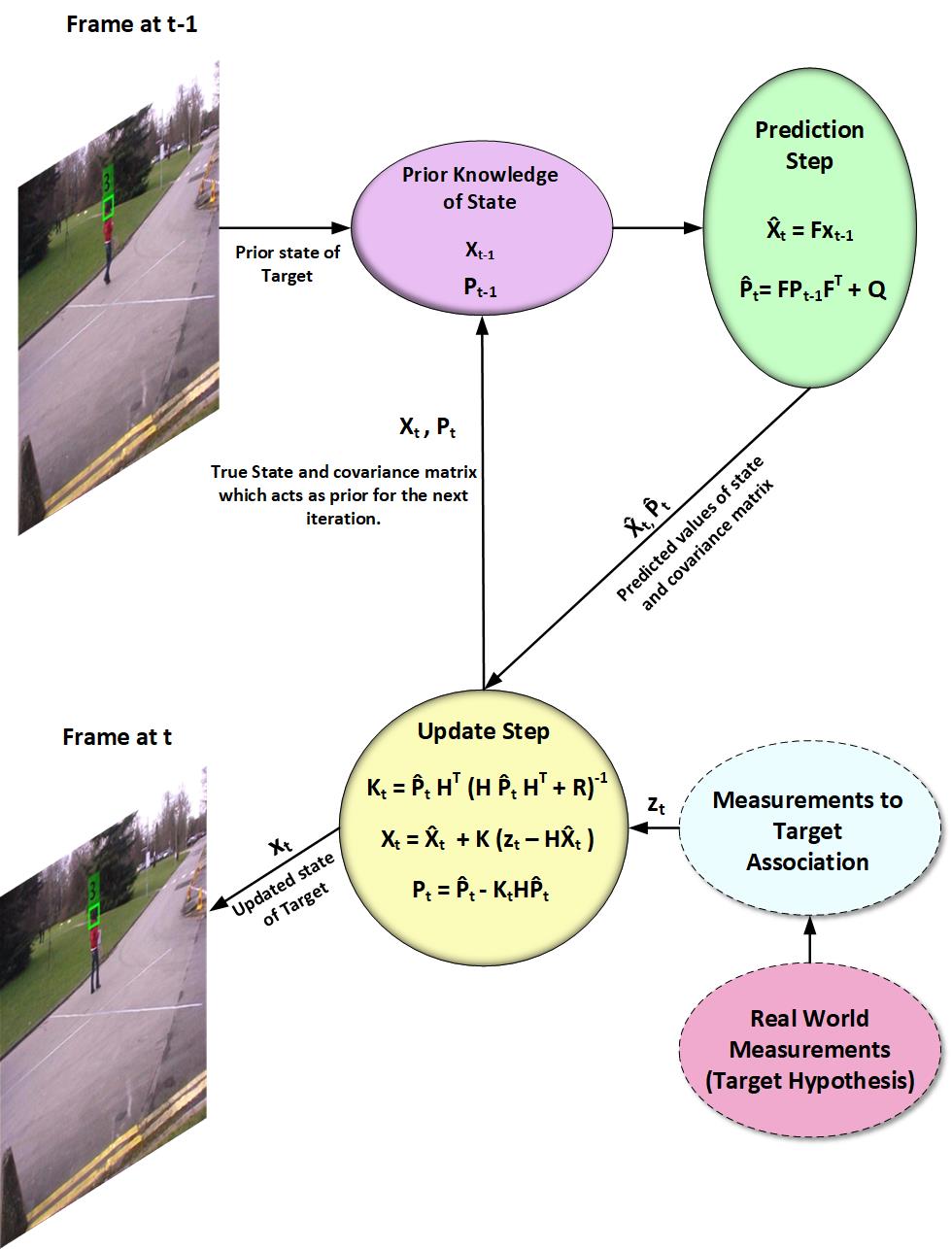}
	\caption{Prediction and the update step of the Kalman filter. It not only keeps track of the target state $x_t$ but also the uncertainty $P_t$ of the state. 
		Initially, a prediction is made for the state of the target using the previously known state and the state transition model. In the update step, the measurement $z_t$ is used to correct the predicted state. 
		In an iterative process, the target is tracked between two consecutive frames.}
	\label{KLF}
\end{figure}

\section{Proposed Approach}
\label{PA}

The proposed approach is based on the tracking-by-detection paradigm. Initially, the target hypothesis is generated in every frame. In our case, the target hypothesis is the spatial position of the target head. In theory, any rigid body part can be used as the target hypothesis. But due to the most important and the most visible position of the head, we select it as the key location for tracking. Our tracker is based on the Kalman filter. The block diagram of the Kalman filter is given in Fig. \ref{KLF}. We assumed a smooth and constant velocity model for the targets in the scene. This is a reasonable assumption because once a target appear in the visual scene, it can not disappear abruptly. Similarly, the motion of the target is smooth as long as it stays in the scene. We modeled the target association as a combinatorial optimization problem. Association is important as with every time step, we have $N$ numbers of target hypothesis and $M$ number of tracks. In order to track the targets as accurately as possible, the correct hypothesis should be assigned to the corresponding tracks. Hence, at every time instance, we produce a $N\times M$ matrix and used the Hungarian assignment algorithm to get the correct associations. A brief description of the Kalman Filter and the assignment algorithm is given in section \ref{tracker} and \ref{HA}, respectively.

\iffalse

The block diagram of the proposed model is given in Figure \ref{PL}. Essentially, the network consists of a Siamese network with a contrastive loss function. Siamese neural network is a special kind of neural network that consists of two parallel Convolutional Neural Networks (CNN). The architecture of both the networks are similar and they share common weight. Essentially, one network is the mirrored version of the other. Each network is taking an image patch corresponding to the target of interest. The target patches are generated by the target detector or manually extracted from the image. The networks extract the discriminative feature through it's linear and non-linear layers (convolution, activation, pooling) from the patches and used as the visual representation of the target. Both the visual representation is given to a contrastive loss function module which outputs the similarity score between the two patches. Compared to the classical CNN, where the network learned to classify the inputs into different categories, the Siamese network gives the dissimilarity score between the corresponding inputs. In the next section \ref{NA}, a brief description of the architecture and loss function is given.  

%Siamese networks are a special type of neural network architecture. Instead of a model learning to classify its inputs, the neural networks learns to differentiate between two inputs. It learns the similarity between them.

\fi

\section{Kalman Filter}
\label{tracker}

Kalman filter is an Online filtering algorithm. Its graphical model is similar to a hidden Markov model. However, it assumes that the process $v_{k-1}$ and measurement $n_k$ noises as well as the posterior pdf $p(x_k|z_{1:k})$ are normally distributed (Fig. \ref{KLF}). Moreover, the function $f_k$ and $h_k$ are linear. Based on these assumptions, 
the following state transition and measurement equations are conceived. 

\begin{equation}
x_k = F_k x_{k-1} + v_{k-1}
\end{equation}

\begin{equation}
z_k = H_k x_{k-1} + n_{k}
\end{equation}

The matrix $F_k$ is called the state transition matrix and it helps to predict the current state of the target based on its previous state. Similarly, the matrix $H_k$ associate the observation $z_k$ to the target state $x$. 
The random variables $v_{k-1}$, $n_k$ show the process and measurement noise. They are zero mean, normally distributed with covariance matrices $Q_{k-1}$ and $R_k$ respectively.
A detailed description of the Kalman filter is beyond the scope of this paper. For details, readers may refer to \cite{maskell2001tutorial, bishop2001introduction}. In our problem, we instantiated an instance of the Kalman filter for each target in the visual scene. 

%The architecture is very similar to a classical CNN \cite{krizhevsky2012imagenet, zeiler2014visualizing} and consist of convolution, rectified linear unit, batch normalization, and dropout layers. The size of convolution filter is fixed to be 3$\times$3 which is followed by ReLU, batch normalization and dropout. The dropout ratio is kept to be 0.2. At the end of each CNN, a 1$\times$1 Convolutional layer is placed which results in a fixed size feature vector for the input patch. The feature vector is used as the visual representation of the targets.  

%We will use a standard convolutional neural network architecture. We use batch normalization after each convolution layer, followed by dropout.

%Siamese networks are a special type of neural network architecture. Instead of a model learning to classify its inputs, the neural networks learns to differentiate between two inputs. It learns the similarity between them.

\subsection{Hungarian Algorithm}
\label{HA}

The Hungarian algorithm is a greedy combinatorial optimization algorithm and solves the assignment problem in polynomial time. The tracking problem is modeled as a bipartite graph matching problem where the first set of nodes corresponds to the established trajectories and the seconds set of nodes corresponds to the target hypothesis measurements from the real world. In our case, the measurements correspond to the spatial locations of the head in every frame. The input to the algorithm is a cost matrix with $N$ number of rows and $M$ number of columns. $N$ corresponds to the established trajectories where the $M$ corresponds to the number of measurements at time step $t$. There are a variety of ways to obtain the cost matrix \cite{luo2014multiple}. A detailed description of appearance model based on visual features is illustrated in \cite{ullah2018hand, ullah2019siamese}. In our work, we mainly used the special constraints for calculating the cost matrix. Specially, we measured the Euclidean distance between the targets head location in the current and previous frame and treat it as the cost. The nearer are the targets in the consecutive frames, the smaller will be the cost and most probably, the targets with the least distance correspond to the same targets in the temporal domain. Similarly, the targets that are far from each other in the consecutive frames would yield the highest cost and corresponds to different targets in the temporal domain. Once the cost matrix is obtained, the Hungarian algorithm \cite{jonker1986improving} works in three steps as the following:

\begin{itemize}
	\item \textbf{Row reduction operation:} Find the minimum cost of each row. Then subtract the corresponding minimum from each row entry to ensure at least one zero-entry in each row. 
	\item \textbf{Column reduction operation:} Repeat the same procedure for each column. It will ensure at least on zero entry in each column. 
	\item \textbf{Optimally test:} Find the minimum number of straight lines to cover all the zeros in the cost matrix. If the number of lines covering all the zeros equal to the number of rows and columns, optimality is achieved. However, if the number of lines covering all the zeros is not equal to the number of rows and columns, shift zeros such as to achieve the optimal assignment.  
\end{itemize}

\section{Experiment}
\label{EX}

\begin{figure*}[h]%[t]
	\centering
	\subfigure[Frame 1]{
		\includegraphics[width=.2\textwidth]{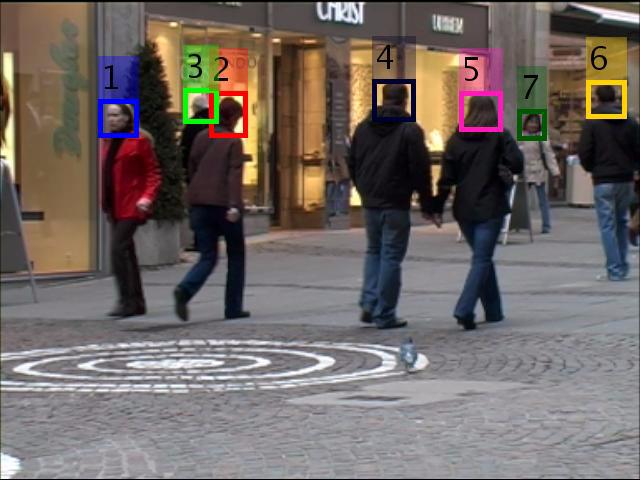}
	}
	\subfigure[Frame 39]{
		\includegraphics[width=.2\textwidth]{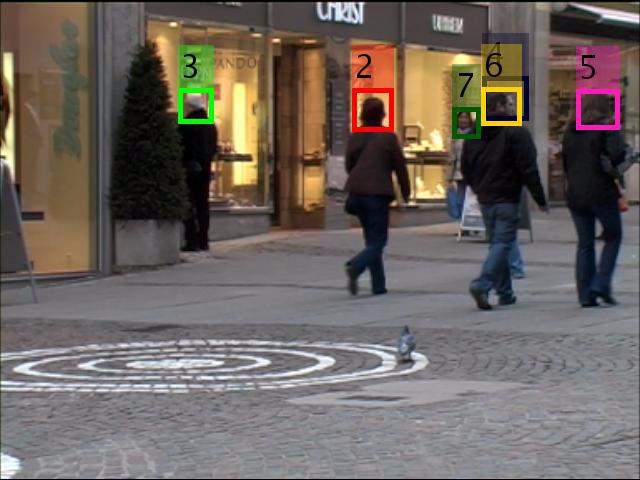}
	}
	\subfigure[Frame 118]{
		\includegraphics[width=.2\textwidth]{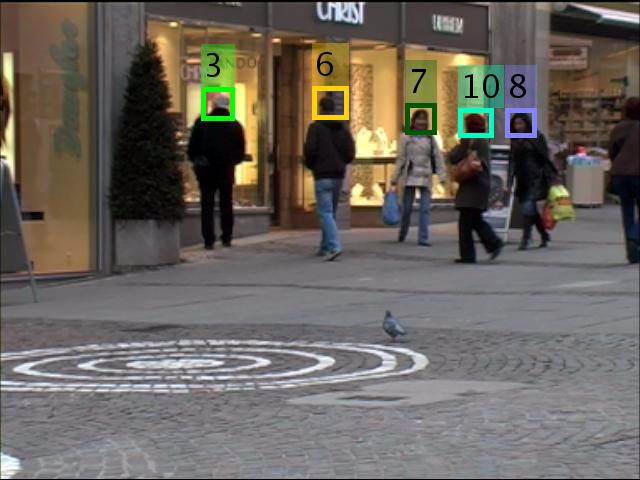}
	}
	\subfigure[Frame 151]{
		\includegraphics[width=.2\textwidth]{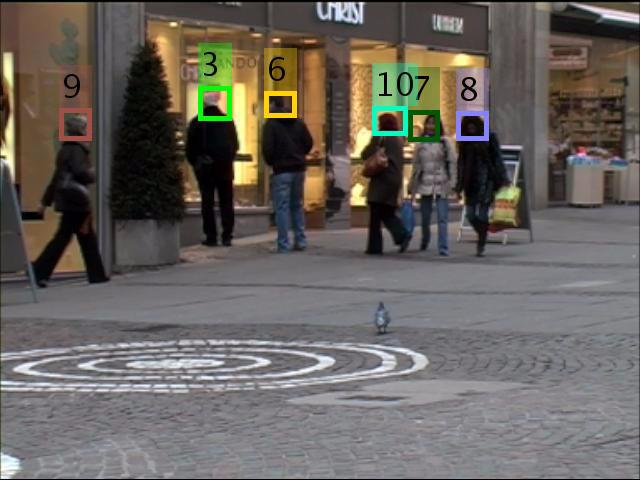}
	}
	
	\subfigure[Frame 5]{
		\includegraphics[width=.2\textwidth]{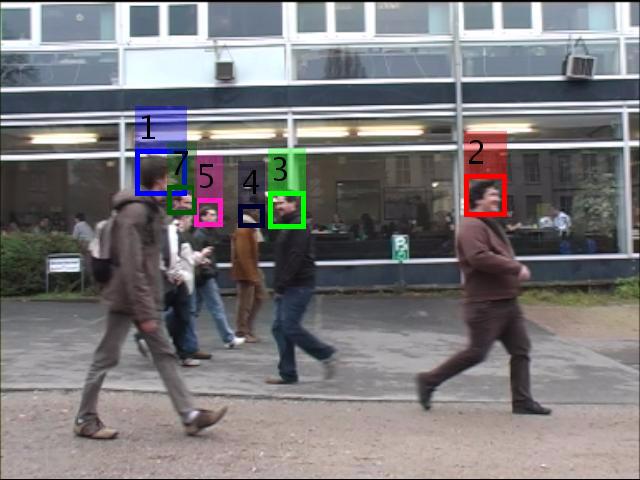}
	}
	\subfigure[Frame 30]{
		\includegraphics[width=.2\textwidth]{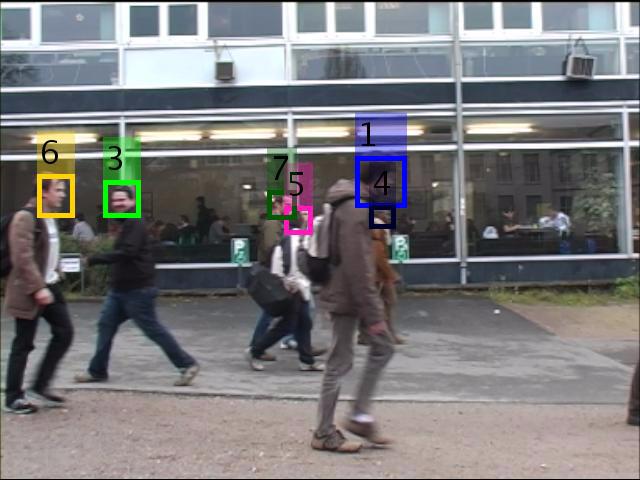}
	}
	\subfigure[Frame 58]{
		\includegraphics[width=.2\textwidth]{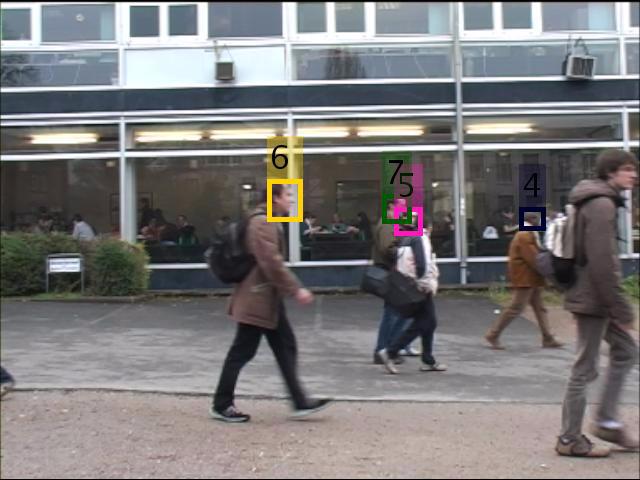}
	}
	\subfigure[Frame 70]{
		\includegraphics[width=.2\textwidth]{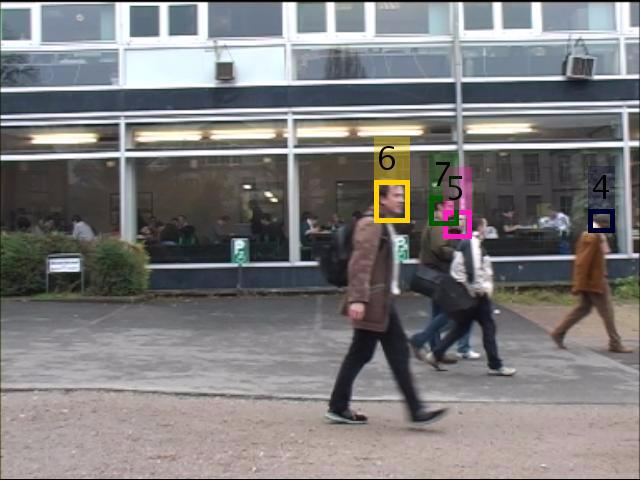}
	}
	
	\subfigure[Frame 2]{
		\includegraphics[width=.2\textwidth]{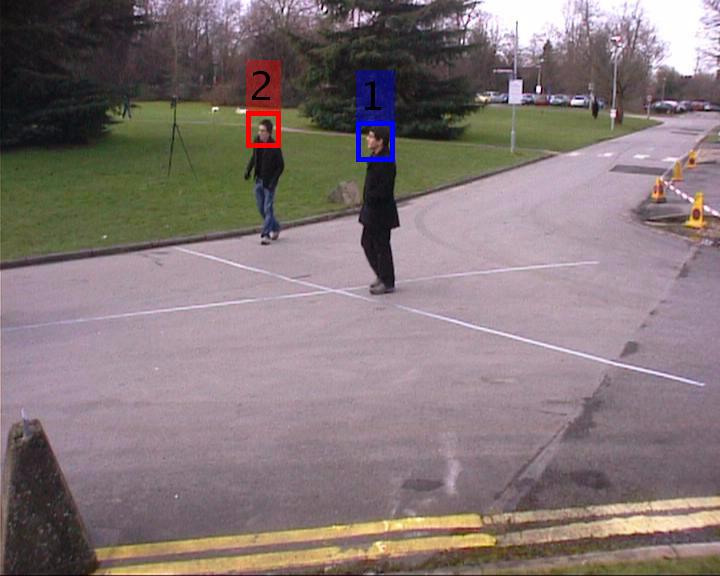}
	}
	\subfigure[Frame 21]{
		\includegraphics[width=.2\textwidth]{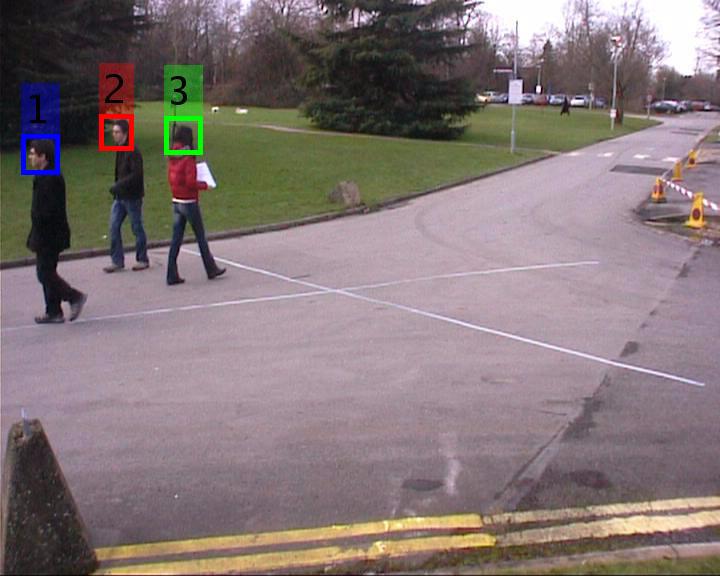}
	}
	\subfigure[Frame 70]{
		\includegraphics[width=.2\textwidth]{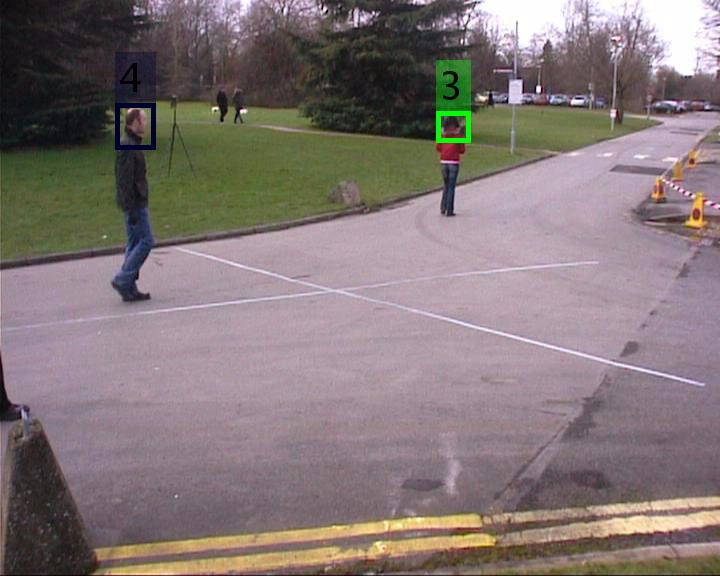}
	}
	\subfigure[Frame 101]{
		\includegraphics[width=.2\textwidth]{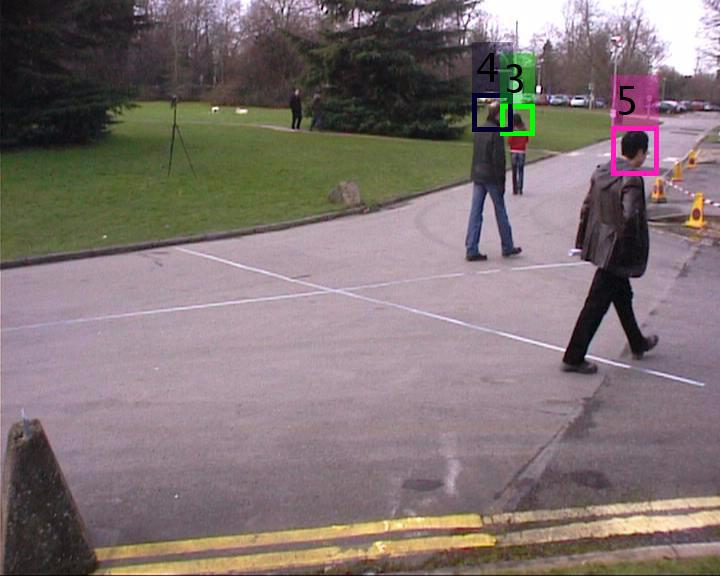}
	}
	
	\subfigure[Frame 4]{
		\includegraphics[width=.2\textwidth]{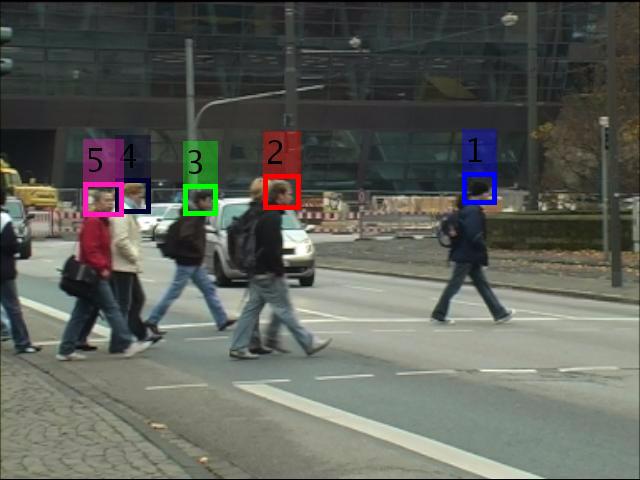}
	}
	\subfigure[Frame 51]{
		\includegraphics[width=.2\textwidth]{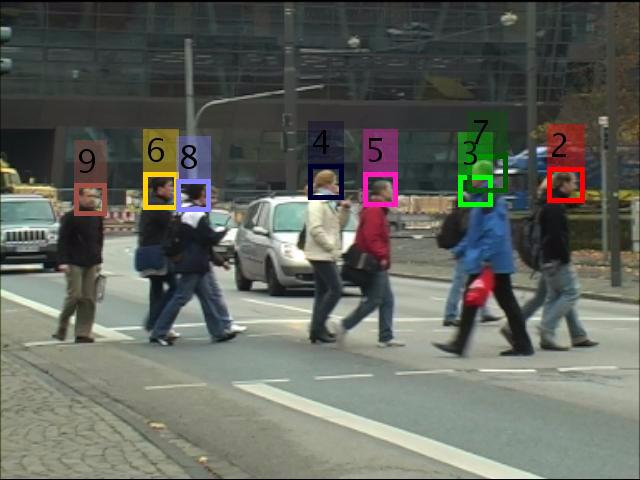}
	}
	\subfigure[Frame 54]{
		\includegraphics[width=.2\textwidth]{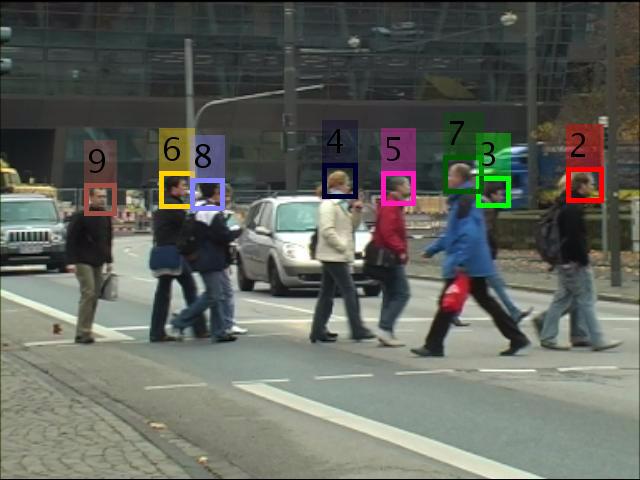}
	}
	\subfigure[Frame 98]{
		\includegraphics[width=.2\textwidth]{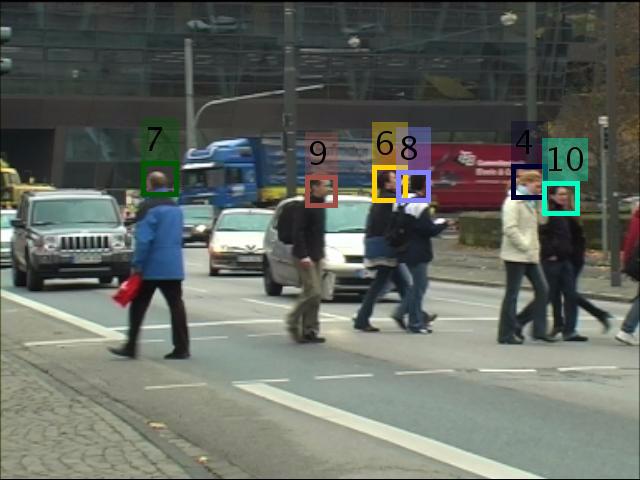}
	}
	
	\caption{Tracking results of the proposed algorithm. Top to left: datasets Stadmitt (a-d), TUD-campus (e-h), Pets2009 (i-l), and TUD-crossing (m-p). Targets are given a unique integer ID.}
	\label{fig:Qresults}
\end{figure*}

The proposed algorithm is implemented in Matlab on a Core i7 system with 16 GB RAM. To ensure real-time performance, we excluded the deep feature based appearance model \cite{ullah2017hierarchical} but it could easily be incorporated in the cost matrix. 
To evaluate the network, we have chosen four datasets \cite{ferryman2009pets2009, riemenschneider2012hough} that are commonly used for pedestrian tracking. It is also worth noticing that all the standard datasets have annotation available but that is for the whole body which is not useful for our case. Therefore, we annotated the datasets to generate the target hypothesis in every frame. The qualitative results of the proposed method are given in Fig. \ref{fig:Qresults}. It is interesting to observe that the head based tracking works well when the heads are not covered with anything. Additionally, due to the most visible part of the body, it also helps in accurate tracking for the partially occluded regions. The proposed algorithm fails when the people use an umbrella or cover the head with an opaque material. However, in the majority of surveillance scenarios where the head is visible, the proposed algorithm works well.

\section{Conclusion}
\label{CON}

We proposed a multi-target tracking algorithm of tracking the heads of multiple humans in the visual scene. The tracking-by-detection paradigm is followed where the spatial locations of the head are generated in every frame and a combinatorial optimization is used to establish the association between the corresponding targets. Especially, the Hungarian algorithm is used to associate the head of the corresponding targets in the consecutive frames in a greedy fashion. In the future, we are aiming to extend our approach to other Keypoint of the body part and rather than tracking the head of a person, track different body parts. Tracking the individual body parts would be a direction for the research for pose estimation and high-level behavior analysis.

{\small
\bibliographystyle{IEEEbib}
\bibliography{CleanNew}
%\bibliography{strings,refs,external}
}
\iffalse
\begin{figure}[!hb]
	\includegraphics[width=0.2\columnwidth]{logo.png}
	\caption{IS\&T logo.}
	\label{Figure:logo}
\end{figure}

\fi
\end{document}